\documentclass{article}
\usepackage{ijcai25}

\usepackage{times}
\usepackage{soul}
\usepackage{url}
\usepackage[hidelinks]{hyperref}
\usepackage[utf8]{inputenc}
\usepackage[small]{caption}
\usepackage{graphicx}
\usepackage{amsmath}
\usepackage{amsthm}
\usepackage{booktabs}
\usepackage{algorithm}
\usepackage{algorithmic}
\usepackage[switch]{lineno}

\usepackage{amssymb}
\usepackage{multirow}
\usepackage{pifont}
\usepackage{dsfont}
\usepackage{subcaption}
\usepackage{newfloat}
\usepackage{listings}
\usepackage{array}
\usepackage{booktabs} 
\usepackage{markdown}
\usepackage{multirow}
\usepackage[table,xcdraw,dvipsnames]{xcolor}

\newtheorem{theorem}{Theorem}
\newtheorem{definition}{Definition}

\newtheorem{lemma}[theorem]{Lemma}
\newtheorem{proposition}{Proposition}

\title{ShortcutProbe: Probing Prediction Shortcuts for Learning Robust Models}

\author{
Guangtao Zheng
\and
Wenqian Ye\And
Aidong Zhang
\affiliations
University of Virginia
\emails
\{gz5hp, pvc7hs, aidong\}@virginia.edu
}

\begin{document}

\maketitle

\begin{abstract}
Deep learning models often achieve high performance by inadvertently learning spurious correlations between targets and non-essential features. For example, an image classifier may identify an object via its background that spuriously correlates with it. This prediction behavior, known as spurious bias, severely degrades model performance on data that lacks the learned spurious correlations. Existing methods on spurious bias mitigation typically require a variety of data groups with spurious correlation annotations called group labels. However, group labels require costly human annotations and often fail to capture subtle spurious biases such as relying on specific pixels for predictions. In this paper, we propose a novel post hoc spurious bias mitigation framework without requiring group labels. Our framework, termed ShortcutProbe, identifies prediction shortcuts that reflect potential non-robustness in predictions in a given model's latent space. The model is then retrained to be invariant to the identified prediction shortcuts for improved robustness. We theoretically analyze the effectiveness of the framework and empirically demonstrate that it is an efficient and practical tool for improving a model's robustness to spurious bias on diverse datasets. 
\end{abstract}

\section{Introduction}
Deep learning models have shown remarkable performance across domains, but this success is often achieved by exploiting spurious correlations ~\cite{sagawadistributionally,liu2021just,yang2023change,labonte2024towards,ye2024spurious,zheng2024benchmarking} between spurious attributes or shortcut features \cite{geirhos2020shortcut} and targets. For example, models have been found to use correlations between textures and image classes~\cite{geirhos2018imagenettrained} for object recognition instead of focusing on defining features of objects. This issue becomes even more problematic in high-stakes domains like healthcare. For instance, models predicting pneumonia were shown to rely on correlations between metal tokens in chest X-ray scans from different hospitals and the disease’s detection outcomes~\cite{zech2018variable}, rather than the pathological features of pneumonia itself. The tendency of using spurious correlations is referred to as \textit{spurious bias}. Models with spurious bias often fail to generalize on data groups lacking the learned spurious correlations, leading to significant performance degradation and non-robust behaviors across different data groups. This robustness issue can have severe social consequences, especially  in critical applications.

Mitigating spurious bias is crucial for robust generalization across data groups with varying spurious correlations. Existing methods on mitigating spurious bias \cite{sagawadistributionally,kirichenko2022last} rely on group labels. Group labels represent spurious correlations with class labels and spurious attributes. For example, \texttt{(waterbirds, water)} \cite{sagawadistributionally} represents a spurious correlation between waterbirds and water backgrounds in the images of waterbirds with water backgrounds. Using group labels specifies explicitly the spurious correlations that a model should avoid. However, obtaining group labels requires expert knowledge and labor-intensive annotation efforts. Moreover, group labels fail to capture subtle spurious biases, such as using certain pixels in images for predictions.

In this paper, we propose a post hoc approach that can automatically detect and mitigate potential spurious biases in a model rather than relying on group labels. Our key innovation is reframing the task of identifying and mitigating spurious biases as detecting and leveraging \textit{prediction shortcuts} in the model's latent space.  Prediction shortcuts are latent features derived from input embeddings and predominantly contribute to producing the same prediction outcome across different classes. In essence, prediction shortcuts represent non-defining features of certain classes that the model heavily uses for predictions. By operating in the model's latent space, our approach leverages the expressiveness of latent embeddings, enabling direct identification of spurious biases across diverse input formats without requiring group labels.  

We present our post hoc approach as a novel framework called \textit{ShortcutProbe}, which first identifies prediction shortcuts in a given model and  leverages them to guide model retraining for spurious bias mitigation. ShortcutProbe utilizes a probe set without group labels, typically containing a diverse mix of features, to uncover potential prediction shortcuts. These  shortcuts are identified as latent features extracted from sample embeddings belonging to different classes but producing the same prediction outcome. By optimizing these features to maximize the model's confidence in their corresponding predictions, ShortcutProbe effectively encodes spurious attributes in non-generalizable prediction shortcuts that the model overly relies on for predictions.

With the identified prediction shortcuts, ShortcutProbe mitigates spurious biases by retraining the model to be invariant to these shortcuts, as they are irrelevant to true prediction targets. This invariance is achieved by applying regularization during retraining, which ensures that the identified prediction shortcuts no longer contribute to the model's predictions of the true targets.

We theoretically demonstrate that when the spurious attributes in the training data are new to the model as reflected by the high prediction loss, the tendency of using spurious attributes for predictions is high after training on the data. Our method aims to revert the process of learning spurious attributes by retraining the model so that the learned spurious attributes induce high prediction losses, effectively unlearning the spurious attributes and reducing the influence of spurious correlations.

In summary, our contributions are as follows:
\begin{itemize}

\item We introduce ShortcutProbe, a novel post hoc framework for mitigating spurious bias without requiring group labels. ShortcutProbe identifies prediction shortcuts and leverages them as a form of regularization for training robust models. 

 \item We provide a theoretical analysis revealing that our spurious bias mitigation approach effectively unlearns spurious attributes.

\item Through extensive experiments, we show that our method successfully trains models robust to spurious biases without prior knowledge about these biases.

\end{itemize}

\section{Related Works}
\paragraph{Shortcuts and spurious bias.} Shortcuts are decision rules that perform well on standard benchmarks but fail to transfer to more challenging testing conditions~\cite{geirhos2020shortcut}, such as data with distributional shifts. Spurious correlations describe superficial associations between spurious (non-essential) features and targets in data, and they can be used by machine learning models as shortcuts~\cite{geirhos2018imagenettrained}. This shortcut-learning phenomenon in machine learning models results in spurious bias, the tendency to use spurious correlations in data for predictions. Shortcut features are spurious attributes that are used by models in predictions. Accurately identifying shortcut features involves domain knowledge and model interpretation techniques. In our work, we circumvent this by identifying prediction shortcuts that represent \textit{potential} shortcut features in a model's latent space, making our method a general and useful method for spurious bias detection and mitigation.

\paragraph{Spurious bias mitigation with group labels.} Group labels in training data indicate the presence of spurious correlations across different subsets of the data.  Exploiting group labels during training helps mitigate the reliance on the specified spurious correlations. Existing approaches for spurious bias mitigation utilize group labels to balance data distributions during training \cite{cui2019class,he2009learning,byrd2019effect}, to formulate a distributionally robust optimization objective \cite{sagawa2019distributionally}, or to progressively expand group-balanced training data \cite{deng2023robust}. Although these methods achieve remarkable success in spurious bias mitigation, the reliance on training group labels becomes a barrier in practice, as obtaining such labels often requires domain knowledge and labor-intensive annotation efforts. Our method does not require group labels during training; instead, it automatically detects prediction shortcuts that encode potential spurious biases the model has developed in training.

\paragraph{Reducing reliance on group labels.} To alleviate the dependency on training group labels, existing approaches propose inferring training group labels through various means, such as identifying misclassified samples~\cite{liu2021just}, clustering hidden representations~\cite{pmlr-v162-zhang22z}, employing invariant learning techniques~\cite{creager2021environment}, or training group label estimators using a few group-labeled samples~\cite{nam2022spread}. Nevertheless, validation group labels are still needed to specify which biases to address. Recent works~\cite{zheng2024learning,zheng2024spuriousness} use vision-language models to infer group labels. Last-layer retraining methods \cite{kirichenko2022last,labonte2024towards} leverage group-balanced validation data to fine-tune the last layer of a model. While our approach also involves retraining the last layer, it does not require group labels. Instead, we introduce a novel strategy that utilizes identified prediction shortcuts during retraining, leading to improved model robustness.

\section{Preliminary}
A \textit{spurious correlation} is the correlation between a  \textit{spurious attribute} present in the training samples and a prediction target. For example, the class \texttt{waterbird} and the attribute \texttt{water background} might form a spurious correlation in the images of \texttt{waterbird}, where some of them have water backgrounds, e.g., pond or river, and some do not. Spurious attributes are not truly predictive of the targets. A group label,  e.g., (\texttt{waterbird}, \texttt{water background}), consists of a prediction target and a spurious attribute.

\begin{figure*}[t]
    \centering
    \includegraphics[width=0.95\linewidth]{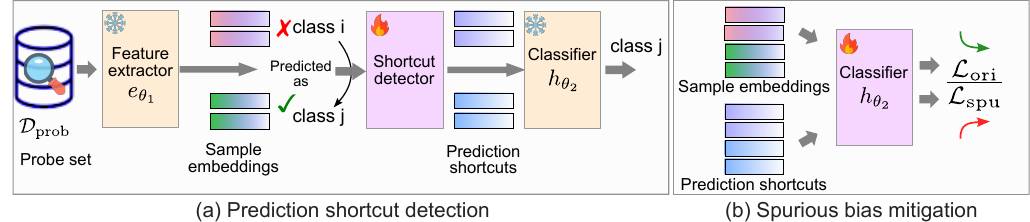}
    \caption{Illustration of ShortcutProbe. (a) The framework uses a set of probe data $\mathcal{D}_{\text{prob}}$ to identify prediction shortcuts by learning a shortcut detector to extract similar features from samples of different classes $i$ and $j$ that are all predicted as the same class $j$. Feature extractor $e_{\theta_1}$ and classifier $h_{\theta_2}$ are frozen during this stage. (b) ShortcutProbe then retrains the classifier with the probe data (the loss of the probe data $\mathcal{L}_{\text{ori}}$) while using the identified prediction shortcuts as regularization (the loss of the prediction shortcuts $\mathcal{L}_{\text{spu}}$).}
    \label{fig:method-overview}
\end{figure*}

Consider a model $f_\theta:\mathcal{X}\rightarrow\mathbb{R}^{|\mathcal{Y}|}$ with parameter $\theta$ trained on a training dataset $\mathcal{D}_{\text{tr}}=\{(x_i,y_i)\}_{i=1}^N$ with $N$ sample-label pairs, where $x_i\in\mathcal{X}$ denotes a sample in the input set $\mathcal{X}$, $y_i\in\mathcal{Y}$ denotes a label in the label set $\mathcal{Y}$, and $|\cdot|$ denotes the size of a set. The model $f_\theta=e_{\theta_1}\circ h_{\theta_2}$ can be considered as a feature extractor $e_{\theta_1}:\mathcal{X}\rightarrow \mathbb{R}^D$ followed by a classifier $h_{\theta_2}:\mathbb{R}^D\rightarrow \mathbb{R}^{|\mathcal{Y}|}$, where $\theta = \theta_1 \cup \theta_2$, $h_{\theta_2}$ is the last layer of the model, and $D$ denotes the number of dimensions. 

Due to the existence of spurious attributes in $\mathcal{D}_{\text{tr}}$, the model can exploit them for predictions, such as recognizing waterbirds by detecting the existence of water backgrounds \cite{sagawa2019distributionally}. 
This presents a challenge: It is hard to determine whether a high-performing model is truly robust or simply ``right for the wrong reasons", i.e., relying on spurious attributes.  Although models with spurious biases typically exhibit degraded performance when the learned spurious attributes are absent from input data, e.g., a waterbird on a land background, it remains challenging to identify the specific spurious attributes encoded by the model without group labels, which hinders the development of effective spurious bias mitigation strategies.

\section{Methodology}
\subsection{Method Overview}
We introduce \textit{ShortcutProbe}, a post hoc framework that automatically detects and mitigates prediction shortcuts without requiring group labels to specify which spurious biases to address. The framework comprises two key steps: (1) \textbf{Prediction shortcut detection}, where a probe set is used to identify prediction shortcuts, and (2) \textbf{Spurious bias mitigation}, where the identified shortcuts are used in retraining the model to mitigate spurious biases in the model.

We provide an overview of our framework in Fig.~\ref{fig:method-overview}. The process begins with a set of probe data $\mathcal{D}_{\text{prob}}$, which typically contains samples with various spurious attributes that reflect a model's non-robustness to spurious correlations. These samples are mapped to the latent embedding space of the model $f_{\theta}$ through its feature extractor $e_{\theta_1}$, allowing us to model \textit{any} prediction shortcuts the model might use for predictions. This strategy bypasses the need to explicitly define spurious correlations through group labels, a task that is especially challenging for subtle features, such as specific pixels in images. More concretely,
as shown in Fig.~\ref{fig:method-overview}(a), we first train a shortcut detector that extracts potential prediction shortcuts from samples of different classes but having the same prediction. The identified prediction shortcuts encode spurious attributes shared across classes and capture the model's non-robustness across different data groups.  Next, in Fig.~\ref{fig:method-overview}(b), the model is retrained with $\mathcal{D}_{\text{prob}}$ to mitigate spurious biases by unlearning the identified prediction shortcuts.

\subsection{Prediction Shortcut Detection}\label{sec:shortcut-detection}
Given a model $f_{\theta}$ and a probe set $\mathcal{D}_{\text{prob}}$, we aim to detect prediction shortcuts by learning a shortcut detector $g_{\psi}:\mathbb{R}^{D}\rightarrow \mathbb{R}^{D}$ to identify prediction shortcuts from input embeddings. 
Intuitively, prediction shortcuts can be identified from samples of different classes but having the same prediction outcome, an indication that similar features exist in these samples but are irrelevant to classes. 
In other words, prediction shortcuts can be shared among samples from different classes, necessitating a shared representational space to encode diverse prediction shortcuts. We formalize this intuition in the following definition.



\begin{definition}[Prediction shortcuts]\label{def:prediction-shortcut}
Given input sample-label pairs $(x,y)$ and $(x',y')$, where $y\neq y'$, a trained model $f_{\theta}=e_{\theta_1}\circ h_{\theta_2}$, sample embeddings $\mathbf{v}=e_{\theta_1}(x)$ and $\mathbf{v'}=e_{\theta_1}(x')$ for $x$ and $x'$, respectively, a vector space $\mathcal{V}\subset\mathbb{R}^D$ spanned by $K$ base column vectors in $\mathbf{A}\in\mathbb{R}^{D\times K}$, prediction shortcuts $\mathbf{s}_{x}\in\mathcal{V}$ and $\mathbf{s}_{x'}\in\mathcal{V}$  for the two samples satisfy the following conditions:
    \begin{itemize}
        \item $\text{Pred}\big(h_{\theta_2}(e_{\theta_1}(x)\big)=y$, $\text{Pred}\big(h_{\theta_2}(\mathbf{s}_x)\big)=y$, and
        \item $\text{Pred}\big(h_{\theta_2}(e_{\theta_1}(x')\big)= y$, $\text{Pred}\big(h_{\theta_2}(\mathbf{s}_{x'})\big)=y$,
    \end{itemize}
where $\text{Pred}(f_\theta(x))=\arg\max_{\mathcal{Y}}f_\theta(x)$, $\mathbf{s}_{x}=\mathbf{P}_{\mathbf{A}}\mathbf{v}$, $\mathbf{s}_{x'}=\mathbf{P}_{\mathbf{A}}\mathbf{v}'$, and 
\begin{equation}\label{eq:projection-matrix}
    \mathbf{P}_{\mathbf{A}}=\mathbf{A}(\mathbf{A}^T\mathbf{A})^{-1}\mathbf{A}^T
\end{equation} 
is the projection matrix such that the prediction shortcut $\mathbf{s}_x$ is the best estimate of $\mathbf{v}$ in the vector space $\mathcal{V}$ in the sense that the distance $\|\mathbf{s}_x-\mathbf{v}\|_2^2$ is minimized.
\end{definition}

In Definition \ref{def:prediction-shortcut}, we define a prediction shortcut as a projection of a sample embedding. It exists in the vector space $\mathcal{V}$ shared by samples of different classes with $K$ degrees of freedom. 
Here, $K$ governs the complexity of the vector space representing prediction shortcuts. A smaller $K$ results in a less expressive vector space that may fail to adequately capture prediction shortcuts. Conversely, a larger $K$ provides greater flexibility in representing prediction shortcuts but may encode irrelevant information. The optimal value of $K$ depends on the complexity of the probe data; values that are too small or too large can impede learning and lead to suboptimal performance. We treat $K$ as a tunable hyperparameter.

By representing prediction shortcuts as vectors, we can in theory capture any spurious bias, even the intricate ones. For instance, a vector $\mathbf{s}$ might correspond to features of water backgrounds that are used to predict waterbirds in any image with water backgrounds, revealing a spurious bias in the model. Alternatively, $\mathbf{s}$ could represent a specific feature corresponding to certain pixels in input images, capturing the prediction shortcut based on low-level pixel values---an aspect that is challenging to articulate through group labels.

\paragraph{Learning the shortcut detector.}  Based on the definition of the prediction shortcut, we design the shortcut detector $g_{\psi}$ as a function that implements the projection operation defined in Eq.~\eqref{eq:projection-matrix} with the learnable parameter $\psi=\mathbf{A}\in\mathbb{R}^{D\times K}$, that is for a sample embedding $\mathbf{v} \in \mathbb{R}^D$, $g_{\psi}(\mathbf{v})=\mathbf{P}_{\mathbf{A}}\mathbf{v}$. Learning $g_{\psi}$ essentially learns a shared vector space spanned by $\mathbf{A}$ that could cover prediction shortcuts in samples in the probe set.

To effectively learn $g_{\psi}$, for each class $y$, we first collect samples from the probe set $\mathcal{D}_{\text{prob}}$ to formulate two non-overlapping sets $\mathcal{D}_{\text{cor}}^y$ and $\mathcal{D}_{\text{pre}}^y$, i.e., 
\begin{equation}\label{eq:correct-samples}
    \mathcal{D}_{\text{cor}}^y=\{(x,y)|(x,y)\in\mathcal{D}_{\text{prob}}, \text{Pred}(f_{\theta}(x))=y\},
\end{equation}
and
\begin{equation}\label{eq:predicted-samples}
    \mathcal{D}_{\text{pre}}^y=\{(x',y')|(x',y')\in\mathcal{D}_{\text{prob}}, \text{Pred}(f_{\theta}(x'))= y\neq y'\},
\end{equation}
where $\mathcal{D}_{\text{cor}}^y$ and $\mathcal{D}_{\text{pre}}^y$ contain samples that are correctly and incorrectly predicted as $y$.
The two sets together demonstrate a possibility that certain features shared across classes are incorrectly associated with prediction targets, allowing us to extract these features as potential prediction shortcuts.

Next, we propose the following objective to identify prediction shortcuts:
\begin{equation}\label{eq:shortcut-detector-objective}
    \mathcal{L}_{\text{det}}=\mathop{\mathbb{E}}\limits_{y\in\mathcal{Y}}\mathop{\mathbb{E}}\limits_{(x,y)\in\mathcal{D}^y_{\text{cor}}\cup\mathcal{D}^y_{\text{pre}}}\ell\big(h_{\theta_2}(g_{\psi}(\mathbf{v})),y\big),
\end{equation}
where $\ell:\mathbb{R}^{|\mathcal{Y}|}\times\mathcal{Y}\rightarrow\mathbb{R}$ is the loss function, $\psi=\mathbf{A}$, and $\mathbf{v}=e_{\theta_1}(x)$. To ensure that prediction shortcuts are relevant to the given samples, we regularize the semantic similarity between $g_{\psi}(\mathbf{v})$ and $\mathbf{v}$ as follows,
\begin{equation}\label{eq:shortcut-detector-reg}
    \mathcal{L}_{\text{reg}}=\mathop{\mathbb{E}}\limits_{y\in\mathcal{Y}}\mathop{\mathbb{E}}\limits_{(x,y)\in\mathcal{D}^y_{\text{cor}}\cup \mathcal{D}^y_{\text{pre}}}\|g_{\psi}(\mathbf{v})-\mathbf{v}\|_2^2.
\end{equation}
The overall learning objective for $\psi$ is
\begin{equation}\label{eq:shortcut-objective}
    \psi^*=\arg\min_{\psi}\mathcal{L}_{\text{det}}+\eta\mathcal{L}_{\text{reg}},
\end{equation}
where $\eta>0$ represents the regularization strength.
The objective in Eq.~\eqref{eq:shortcut-objective} aims to encode the properties of prediction shortcuts in Definition \ref{def:prediction-shortcut} into the shortcut detector $g_{\psi}$ while maintaining the relevance of the prediction shortcuts to input samples. Training $g_{\psi}$ is lightweight as there are only $DK$ learnable parameters. With $g_{\psi}$, we can identify multiple prediction shortcuts from samples in $\mathcal{D}_{\text{prob}}$.


\subsection{Spurious Bias Mitigation}\label{sec:spurious-bias-mitigation}
Mitigating spurious biases in a model requires that the spurious attributes captured during training are no longer predictive of the targets. Although identifying spurious attributes can be challenging, the shortcut detector introduced in the previous section identifies prediction shortcuts as potential spurious attributes utilized by the model, providing valuable guidance for addressing spurious biases.

In the following, we formulate an optimization objective that incorporates the above constraint to learn a robust model using the probe set $\mathcal{D}_{\text{prob}}$.
First, a general requirement is that the trained model should produce correct and consistent predictions on training samples. To this end, for each class $y$, we sample from $\mathcal{D}^y_{\text{cor}}$ and $\mathcal{D}_{\text{mis}}^y$, where $\mathcal{D}^y_{\text{cor}}$ is the set of correctly predicted sample-label pairs defined in  Eq.~\eqref{eq:correct-samples}, and $\mathcal{D}_{\text{mis}}^y$ is defined as follows,
\begin{equation}\label{eq:misclassified-samples}
    \mathcal{D}_{\text{mis}}^y=\{(x',y)|(x',y)\in\mathcal{D}_{\text{prob}}, \text{Pred}(f_{\theta}(x'))\neq y\},
\end{equation}
representing misclassified sample-label pairs from the class~$y$. Then, the training objective $\mathcal{L}_{\text{ori}}$ is as follows,
\begin{equation}\label{eq:main-objective}
\begin{aligned}
    \mathcal{L}_{\text{ori}}(\mathcal{D}_{\text{prob}}; \theta) = \mathop{\mathbb{E}}\limits_{y\in\mathcal{Y}}\mathop{\mathbb{E}}\limits_{(x,y)\in\mathcal{D}^y_{\text{cor}}\cup\mathcal{D}^y_{\text{mis}}}\ell(f_\theta(x),y),
\end{aligned}
\end{equation}
which mitigates potential spurious biases by ensuring consistent predictions in $\mathcal{D}^y_{\text{mis}}$ and $\mathcal{D}^y_{\text{cor}}$. 

Moreover, to ensure that the training targets at mitigating the spurious biases in the model,  we further formulate a regularization term using the prediction shortcuts identified by our shortcut detector to guide the training process. Specifically, as the identified prediction shortcuts are not predictive of the targets, we aim to maximize the loss on the prediction shortcuts defined as follows,
\begin{equation}\label{eq:spu-objective}
    \mathcal{L}_{\text{spu}}(\mathcal{D}_{\text{prob}}; \theta) = \mathop{\mathbb{E}}\limits_{y\in\mathcal{Y}}\mathop{\mathbb{E}}\limits_{(x,y)\in\mathcal{D}^y_{\text{cor}}\cup\mathcal{D}^y_{\text{mis}}}\ell(h_{\theta_2}(g_{\psi}(\mathbf{v})),y),
\end{equation}
where $\mathbf{v}=e_{\theta_1}(x)$.

We combine the two loss terms above to formulate the  \textbf{overall training objective} as follows,
\begin{equation}\label{eq:overall-objective}
    \theta_2^*=\arg\min_{\theta_2}\lambda\mathcal{L}_{\text{ori}} / \mathcal{L}_{\text{spu}},
\end{equation}
where $\lambda>0$ is the regularization strength. Here, we retrain only the final classification layer of the model while keeping the feature extractor frozen. This approach significantly reduces computational complexity and allows us to reuse the previously learned sample embeddings. Details of the training algorithm are provided in Appendix.

\subsection{Choice of the Probe Set}
The probe set plays a crucial role in both detecting prediction shortcuts and mitigating spurious biases.  To achieve the full potential of ShortcutProbe, we use a held-out dataset—unseen by the model—to construct the probe set $\mathcal{D}_{\text{prob}}$. This choice is essential because the model may have memorized the training samples, making it difficult to identify prediction shortcuts based on discrepancies in predictions for samples of the same class. Moreover, we select samples with high predication confidence (measured by output entropy) from the held-out dataset to construct $\mathcal{D}_{\text{prob}}$ so that prediction shortcuts can be easily detected from these samples. Details of constructing $\mathcal{D}_{\text{prob}}$ and results on different choices of $\mathcal{D}_{\text{prob}}$ are provided in Section~\ref{sec:experiment}.

\subsection{Theoretical Analysis}
We theoretically demonstrate that minimizing the proposed objective $\mathcal{L}_{\text{ori}}/\mathcal{L}_{\text{reg}}$ effectively unlearns the spurious correlations between spurious attributes and their associated targets captured in the model.
Without loss of generality, we analyze this in the context of a general linear regression setting. Consider an
input sample $x\in\mathcal{X}$, a prediction target $y\in\mathcal{Y}$, and a spurious-only sample $\tilde{x}$ that lacks any defining features in $x$ related to $y$. Let $x_1, \ldots, x_N$ denote $N$ training samples, $\varphi:\mathcal{X}\rightarrow\mathbb{R}^{D}$ be  a generic feature map, and $J_\mathbf{w}(x) = \varphi(x)^T\mathbf{w}$ represent a generalized linear regression model with parameters $\mathbf{w}\in\mathbb{R}^D$.

We denote the correlation between the model output for a spurious-only sample $\tilde{x}$ and a prediction target $y$ as $\rho(J_{\mathbf{w}}(\tilde{x}),y)$. The following lemma \cite{Bombari2024how} gives an upper bound on the correlation. 
\begin{lemma}\label{lemma:spurious-bound}
The correlation between the model output for the spurious-only sample $\tilde{x}$ and the prediction target $y$ is upper bounded as follows:
\begin{equation}
    \rho(J_{\mathbf{w}}(\tilde{x}),y) \leq  \gamma_{\varphi}\sigma_{\mathcal{Y}}\sqrt{\mathcal{R}_{\mathcal{X}}},
\end{equation}
where $\mathcal{R}_{\mathcal{X}}$ is the generalization error, $\sigma_{\mathcal{Y}}$ is the standard deviation of prediction targets, and 
\begin{align}\label{eq:aligment-term}
\gamma_{\varphi} = \mathbb{E}_{\tilde{x},x}\Big[\frac{\varphi(\tilde{x})^T\mathbf{O}\varphi(x)}{\|\mathbf{O}\varphi(x)\|_2^2}\Big],
\end{align}
where $\mathbf{O}=\mathbf{I}-\mathbf{V}^T(\mathbf{V}\mathbf{V}^T)^{-1}\mathbf{V}$ is the orthogonal projection matrix, and $\mathbf{V}=[\varphi(x_1),\ldots,\varphi(x_N)]^T\in\mathbb{R}^{N\times D}$ is the feature matrix.
\end{lemma}
Given that $\mathcal{R}_\mathcal{X}$ and $\sigma_{\mathcal{Y}}$ are independent of $\tilde{x}$, the spurious-only sample $\tilde{x}$ affects the correlation upper bound via the feature alignment term $\gamma_{\varphi}$ between the spurious attribute of $\tilde{x}$ and the original feature of $x$. 
We further interpret this term in the following proposition.

\begin{proposition} The feature alignment term $\gamma_{\varphi}$ is the ratio between the expected prediction error on the spurious sample $\tilde{x}$ and the expected prediction error on the original sample $x$: 
\begin{align}
\gamma_{\varphi}&=\mathbb{E}_{\tilde{x},x}\Big[\frac{\varphi(\tilde{x})^T\mathbf{O}\varphi(x)}{\|\mathbf{O}\varphi(x)\|_2^2}\Big]
=\frac{\mathbb{E}_{\tilde{x}}[\|\mathbf{O}\varphi(\tilde{x})\|_2]}{\mathbb{E}_{x}[\|\mathbf{O}\varphi(x)\|_2]}.\label{eq:feature-alignment}
\end{align}
\end{proposition}
The term $\|\mathbf{O}\varphi(x)\|_2$ in Eq. \eqref{eq:feature-alignment} denotes the error term for the sample $x$, while $\|\mathbf{O}\varphi(\tilde{x})\|_2$ denotes the error term for the spurious sample $\tilde{x}$. Since $x$ and $\tilde{x}$ are independent, the feature alignment term $\gamma_{\varphi}$ is the ratio between the loss on spurious-only samples and the loss on the original samples. We provide a detailed proof in Appendix.  

The prediction shortcuts obtained by our shortcut detector approximate the spurious-only features. Thus, the loss $\mathcal{L}_{\text{spu}}$  approximates the nominator of the regularization term in Eq.~\eqref{eq:feature-alignment}. Moreover,  $\mathcal{L}_{\text{ori}}$ approximates the denominator in Eq.~\eqref{eq:feature-alignment}. Note that Lemma 1 gives the upper bound of the spurious correlation \textit{before} learning it. The objective in Eq.~\eqref{eq:shortcut-detector-objective} trains the shortcut detector to learn the correlation by minimizing $\gamma_{\phi}$. In the mitigation step, the objective in Eq.~\eqref{eq:overall-objective} unlearns the correlation by maximizing $\gamma_{\phi}$.

\section{Experiments}\label{sec:experiment}
\subsection{Datasets}

\paragraph{Image datasets.} \textbf{Waterbirds} 
\cite{sagawa2019distributionally} contains waterbird and landbird classes selected from the CUB dataset \cite{WelinderEtal2010}. The two bird classes are mixed with water and land backgrounds  from the Places dataset \cite{zhou2017places}.
\textbf{CelebA} \cite{liu2015deep} is a large-scale image dataset of celebrity faces. Images showing two hair colors, non-blond and blond, are spuriously correlated with gender. 
\textbf{CheXpert} \cite{irvin2019chexpert} is a chest X-ray dataset containing six spurious attributes from the combination of race (White, Black, Other) and gender (Male, Female). Two diagnose results, i.e., ``No Finding" (positive) and ``Finding" (negative) are the labels.
\textbf{ImageNet-9} \cite{ilyas2019adversarial} is a subset of ImageNet \cite{imagenet} and contains nine super-classes. It is known to have correlations between object classes and image textures. We prepared the training and validation data as in \cite{kim2022learning} and \cite{bahng2020learning}. \textbf{ImageNet-A} \cite{hendrycks2021natural} is a dataset of real-world images, adversarially curated to test the limits of classifiers such as ResNet-50. While these images are from standard ImageNet classes \cite{imagenet}, they are often misclassified in multiple models. We used this dataset to test the robustness of a classifier after training it on ImageNet-9.
\noindent \textbf{NICO} \cite{he2021towards} is designed for out-of-distribution image classification, simulating real-world scenarios where testing distributions differ from training ones. It labels images with both main concepts (e.g., cat) and contexts (e.g., at home). We used the Animal super-class in NICO and followed the setting in \cite{bai2021decaug,pmlrv202tiwari23a} for data preparation.

\begin{table*}[t]
\centering
\resizebox{\linewidth}{!}{%
\begin{tabular}{cccccccccc}
\toprule
\multirow{2}{*}{Method}  & \multicolumn{3}{c}{Waterbirds} & \multicolumn{3}{c}{CelebA} & \multicolumn{3}{c}{CheXpert} \\ \cmidrule(lr){2-10}                   & WGA ($\uparrow$)         &Average ($\uparrow$)    & Gap ($\downarrow$)    & WGA ($\uparrow$)       & Average ($\uparrow$)& Gap ($\downarrow$)     & WGA ($\uparrow$)        & Average ($\uparrow$) & Gap ($\downarrow$)     \\ \midrule
ERM    \cite{vapnik1999overview}                                     &     80.3$_{\pm3.1}$         &     93.3$_{\pm0.4}$  & 13.0          &    45.6$_{\pm2.9}$        &     95.2$_{\pm0.2}$   & 49.6       &    22.0$_{\pm1.6}$         &     90.8$_{\pm0.1}$   &  68.8       \\ \midrule
JTT  \cite{liu2021just}                                   &   86.7$_{\pm1.0}$            &    93.3$_{\pm0.2}$   & 6.6          &     40.6$_{\pm1.2}$        &      88.6$_{\pm0.2}$         &   48.0 &     60.4$_{\pm4.9}$    &       75.2$_{\pm0.8}$ & 14.8        \\
DFR \cite{kirichenko2022last}                    &               90.3$_{\pm2.1}$         &    95.0$_{\pm1.3}$  & 4.7     &  72.2$_{\pm2.0}$         &    \textbf{92.9}$_{\pm0.1}$  & 20.7         &   72.7$_{\pm1.5}$        &      78.7$_{\pm0.4}$   & 6.0   \\
AFR  \cite{qiu2023simple}                    &             88.7$_{\pm4.2}$       &      95.0$_{\pm1.0}$  & 6.3        &     77.8$_{\pm1.5}$          &    91.0$_{\pm0.4}$  & 13.2       &        72.4$_{\pm2.0}$    &   76.8$_{\pm1.1}$    & 4.4         \\
\cellcolor[HTML]{EFEFEF} \textbf{ShortcutProbe (Ours)}                                    &  \cellcolor[HTML]{EFEFEF}   \textbf{90.8}$_{\pm0.6}$         &   \cellcolor[HTML]{EFEFEF}    \textbf{95.0}$_{\pm0.3}$ & \cellcolor[HTML]{EFEFEF}\textbf{4.2}      &  \cellcolor[HTML]{EFEFEF}   \textbf{83.4}$_{\pm0.9}$ &  \cellcolor[HTML]{EFEFEF}     91.4$_{\pm0.1}$     & \cellcolor[HTML]{EFEFEF}  \textbf{8.0}& \cellcolor[HTML]{EFEFEF} \textbf{75.0}$_{\pm0.7}$         &  \textbf{79.0}$_{\pm0.2}$  \cellcolor[HTML]{EFEFEF}  & \cellcolor[HTML]{EFEFEF} \textbf{4.0}          \\ \bottomrule
\end{tabular}%
}
\caption{Comparison of worst-group accuracy (WGA) and average accuracy (\%) with baseline methods on the Waterbirds, CelebA, and CheXpert datasets. The best results are highlighted in \textbf{boldface}. All bias mitigation methods use the same half of the validation set.}\label{tab:image-datasets}
\end{table*}

\begin{table*}[t]
\centering
\resizebox{0.8\linewidth}{!}{%
\begin{tabular}{ccccccc}
\toprule
\multirow{2}{*}{Method}  & \multicolumn{3}{c}{MultiNLI} & \multicolumn{3}{c}{CivilComments} \\ \cmidrule(lr){2-7} 
                              & WGA ($\uparrow$)        & Average ($\uparrow$) & Gap ($\downarrow$)       & WGA ($\uparrow$)         & Average   ($\uparrow$)  & Gap ($\downarrow$)      \\ \midrule
ERM \cite{vapnik1999overview}                                       &   67.0$_{\pm0.4}$      &    82.2$_{\pm0.2}$  & 15.2           &         58.5$_{\pm1.3}$     &         92.2$_{\pm0.1}$  & 33.7    \\\midrule
JTT   \cite{liu2021just}                         &  71.6$_{\pm0.8}$           &   80.7$_{\pm0.4}$    & 9.1         &    68.3$_{\pm0.9}$           &     \textbf{89.0}$_{\pm0.3}$      & 20.7        \\
DFR     \cite{kirichenko2022last}                      &     72.6$_{\pm1.7}$       &    81.8$_{\pm0.4}$  & 9.2         &       76.6$_{\pm0.8}$        &        85.8$_{\pm0.5}$   & 9.2        \\
AFR      \cite{qiu2023simple}                      &      66.6$_{\pm0.3}$       &      82.2$_{\pm0.2}$   & 15.6     &    74.6$_{\pm 5.1}$         &    84.7$_{\pm2.5}$    & 10.1          \\
\cellcolor[HTML]{EFEFEF} \textbf{ShortcutProbe (Ours)}                                   &   \cellcolor[HTML]{EFEFEF}   \textbf{74.3}$_{\pm0.7}$       &  \cellcolor[HTML]{EFEFEF}  \textbf{82.6}$_{\pm0.3}$  & \cellcolor[HTML]{EFEFEF}   \textbf{8.3}& \cellcolor[HTML]{EFEFEF}   \textbf{79.9}$_{\pm0.6}$         &   \cellcolor[HTML]{EFEFEF}   88.5$_{\pm0.2}$ &\cellcolor[HTML]{EFEFEF}  \textbf{8.6}            \\ \bottomrule
\end{tabular}%
}
\caption{Comparison of worst-group accuracy (WGA) and average accuracy (\%)  with baseline methods on the MultiNLI and Civilcomments datasets.  Best results are highlighted in \textbf{boldface}. All bias mitigation methods use the same half of the validation set.}\label{tab:text-datasets}
\end{table*}

\paragraph{Text datasets.}
\textbf{MultiNLI}~\cite{williams2017broad} is a text classification dataset with 3 classes: neutral, contradiction, and entailment, representing the natural language inference relationship between a premise and a hypothesis. The spurious feature is the presence of negation, which is highly correlated with the contradiction label. 
\textbf{CivilComments}~\cite{borkan2019nuanced} is a binary text classification dataset aimed at predicting whether a comment contains toxic language. Spurious features involve references to eight demographic identities: male, female, LGBTQ, Christian, Muslim, other religions, Black, and White. 
 
\subsection{Experimental Setup}\label{sec:expr-setup}

\paragraph{Constructing the probe set.} From the chosen data source, such as the training or validation set, we sorted the samples within each class by their prediction losses and divided them into two equal halves: a high-loss set and a low-loss set. This approach approximates the incorrectly and correctly predicted samples, respectively, while ensuring that the incorrectly predicted set is non-empty, even when all samples are correctly classified. Within each set, we then selected the top $r$\% of samples with the highest prediction confidence (i.e., those with the lowest output entropy). 

\paragraph{Training details.} We first trained a base model initialized with pretrained weights using empirical risk minimization  (ERM) on the training dataset. Then, we retrained the model on half of the validation set using various bias mitigation methods. For our method, we first constructed the probe set using the same half of the validation set and used the probe set for shortcut detection and mitigation. The remaining half of the validation set was used for model selection and hyperparameter tuning.
For experiments on the Waterbirds, CelebA, and CheXpert datasets, we used ResNet-50 as the backbone network, and we used ResNet-18 on the ImageNet-9/A and NICO datasets to ensure a fair comparison with baseline methods. For text datasets, we used a pretrained BERT model \cite{kenton2019bert}. We trained models using each method three times with different random seeds and reported the average results as well as their standard deviations. Detailed training settings are provided in Appendix\footnote{Code is available at \url{https://github.com/gtzheng/ShortcutProbe}.}.

\paragraph{Metrics.} Without group labels, we used the worst-class accuracy \cite{yang2023change} for model selection, which is defined as the worst per-class accuracy on an evaluation set. For performance evaluation on the Waterbirds, CelebA, CheXpert, MultiNLI, and Civilcomments datasets, we adopted the widely accepted metric, \textit{worst-group accuracy}, which is the lowest accuracy across multiple groups of the test set with each group containing a certain spurious correlation. We also calculated the \textit{accuracy gap} defined as the standard average accuracy minus the worst-group accuracy to measure the degree of a classifier's spurious biases. A high worst-group accuracy with a low accuracy gap indicates that the classifier is robust to spurious biases and can fairly predict samples from different groups. We adopted \textit{average accuracy} for the evaluations on the NICO, ImageNet-9, and ImageNet-A datasets as these datasets are specifically constructed to evaluate the robustness against distributional shifts. 


\subsection{Analysis of Probe Set}
Our method relies on a probe set for detecting prediction shortcuts and mitigating spurious bias. A good probe set can be used to effectively reveal and mitigate spurious biases in a model, such as those curated with group labels \cite{sagawa2019distributionally}.  
Here, we show that our method, ShortcutProbe, performs effectively without group labels when a probe set is carefully selected from \textit{readily available sources}, such as the training data and held-out validation data.

To demonstrate, we constructed a probe set as described in Section~\ref{sec:expr-setup}, using the training set or half of the validation set (with the other half reserved for model selection) as the data source. For each data source, we varied the proportion $r$\% from 20\% to 100\%. This adjustment created different probe sets, ranging from those containing only samples with highly confident predictions (small $r$) to those including all samples from the selected data source ($r=100$). 

\begin{figure}[t]
    \centering
    \includegraphics[width=\linewidth]{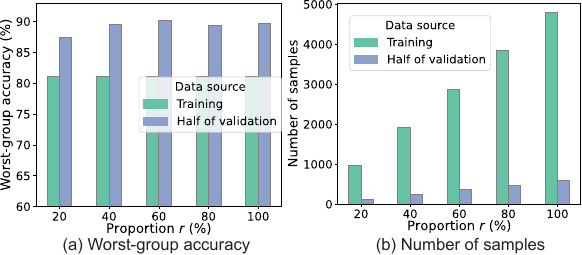}
    \caption{Analyses on different probe sets constructed from the Waterbirds dataset. (a) Worst-group accuracy comparison between models trained with training data and half of the validation data. (b) Numbers of samples in respective probe sets.}
    \label{fig:probe-waterbirds-chexpert}
\end{figure}

Fig.~\ref{fig:probe-waterbirds-chexpert}(a) shows the performance of ShortcutProbe measured by worst-group accuracy (WGA) under different probe sets, while Fig.~\ref{fig:probe-waterbirds-chexpert}(b) presents the sizes of these probe sets. Compared to ERM models, we observe that using the training data for retraining results in minimal improvement on the Waterbirds dataset. This is because most of the training data can be correctly predicted by the model, resulting in a probe set that is not informative for learning prediction shortcuts.

In contrast, by leveraging a relatively small amount of the held-out data compared to the size of the training data, our method demonstrates a significant improvement in robustness to spurious biases. Additionally, our approach benefits most from samples with high prediction confidence, i.e., when $r$ is small. However, as shown in Fig.~\ref{fig:probe-waterbirds-chexpert}(a), setting $r$ too small results in an insufficient number of training samples, leading to suboptimal WGA performance. In the following, unless otherwise specified, we use half of the validation set to construct the probe set and treat $r$ as a tunable hyperparameter.

\subsection{Main Results}
We focus on a challenging and practical setting where group labels are unavailable in both training and validation data. This scenario requires detecting and mitigating spurious biases using only the available data and model in a standard ERM training setup.
As baselines, we selected state-of-the-art last-layer retraining methods DFR \cite{kirichenko2022last} and AFR \cite{qiu2023simple}, as well as JTT \cite{liu2021just}, which retrains the entire model.
For DFR, which typically requires group labels, we used class labels instead. All baseline methods listed in Tables~\ref{tab:image-datasets} and \ref{tab:text-datasets} utilized half of the validation data for training. Similarly, we used the same half of the validation data to construct a probe set, derived as a subset of this portion. The remaining half of the validation data was reserved for model selection.

As shown in Tables~\ref{tab:image-datasets} and \ref{tab:text-datasets}, our method achieves the highest WGA and the smallest accuracy gap between average accuracy and WGA, indicating its effectiveness in balancing performance across different data groups. Methods that achieve high average accuracy, such as DFR on the CelebA dataset, prioritize learning spurious features in the probe set. Although maintaining good predictivity on average, DFR still suffers from prediction shortcuts, as shown by its low WGA. Our method remains effective on larger backbone networks beyond ResNet-50, such
as ResNet-152 and ViT (see Appendix).
Unlike other baseline methods, our method uses only a portion of available data for training, highlighting the effectiveness of the probe set in detecting and mitigating spurious biases. Notably, when we applied the same probe set with baseline methods, this led to degraded performance in both WGA and accuracy gap, underscoring the unique advantages of our approach.

\begin{table}[t]
\centering
\resizebox{0.8\linewidth}{!}{%
\begin{tabular}{lc}
\toprule
Method                                          & Accuracy ($\uparrow$)    \\ \midrule
ERM                                             & 75.9                \\
REx \cite{krueger2021out}                       & 74.3                \\
Group DRO  \cite{sagawa2019distributionally}    & 77.6                \\
JiGen \cite{carlucci2019domain}                 & 85.0                \\
Mixup \cite{zhang2017mixup}                     & 80.3                \\
CNBB  \cite{he2021towards}                      & 78.2                \\
DecAug \cite{bai2021decaug}                     & 85.2                \\ 
SIFER \cite{tiwari2023overcoming}               & 86.2                \\
\cellcolor[HTML]{EFEFEF}\textbf{ShortcutProbe (Ours)}            & 
\cellcolor[HTML]{EFEFEF} \textbf{90.5}$_{\pm0.6}$ \\ \bottomrule
\end{tabular}%
}
\caption{Comparison of average accuracy (\%) on the NICO dataset.}\label{tab:nico}
\end{table}

We further tested the out-of-distribution generalization of our method on the NICO dataset. Images of each class in the test set are associated with an unseen context. Our method achieves the best classification accuracy without using group labels (Table \ref{tab:nico}), demonstrating its effectiveness in mitigating the reliance on contexts.
We present additional results on the ImageNet-9 and ImageNet-A datasets in Appendix to demonstrate our method's effectiveness in combating distributional shifts and the capability of achieving good tradeoff between in-distribution and out-of-distribution performance.

\subsection{Ablation Studies}
\begin{figure}[t]
    \centering
    \includegraphics[width=\linewidth]{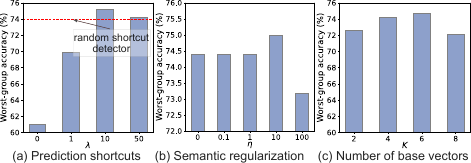}
    \caption{Analyses on how (a) prediction shortcuts as well as their regularization strength $\lambda$, (b) semantic regularization strength $\eta$, and (c) number of base vectors $K$ affect a model's robustness to spurious bias. We report the worst-group accuracy on the CheXpert dataset.}
    \label{fig:ablation}
\end{figure}

\paragraph{Prediction shortcuts.} We evaluated the effectiveness of prediction shortcuts in Fig.~\ref{fig:ablation}(a). We began by using a randomly initialized shortcut detector to optimize the objective in Eq.\eqref{eq:overall-objective}. Additionally, we tested the performance without the spurious bias regularization term $\mathcal{L}_{\text{spu}}$, represented as $\lambda=0$ in Fig.~\ref{fig:ablation}(a), as well as for $\lambda$ values of 1, 10, and 50. Our results indicate that the model performs the worst when prediction shortcuts are not used as regularization. Interestingly, our method still demonstrates effectiveness even when the prediction shortcuts are random. As $\lambda$ increases, the regularization strength decreases. The model achieves optimal performance when an appropriately balanced $\lambda$ is selected.
\paragraph{Semantic regularization.} We analyzed the impact of the semantic regularization strength $\eta$ defined in Eq.~\eqref{eq:shortcut-detector-reg}. Fig.~\ref{fig:ablation}(b) shows that incorporating this regularization with a moderate value of $\eta$ enhances the model's robustness.

\paragraph{Number of base vectors.} The number of base vectors, $K$, determines the representational capacity of the shortcut detector. A value of $K$ that is too small will limit the detector’s ability to identify spurious attributes effectively, while an excessively large $K$ may lead to overfitting on the probe data. Notably, as shown in Fig. \ref{fig:ablation}(c), the optimal value of $K$ is 6, which coincides with the true number of spurious attributes in the CheXpert dataset.


\section{Conclusion}
In this work, we proposed a novel post hoc framework to mitigate spurious biases without requiring group labels. Our approach first learns a shortcut detector in the latent space of a given model via a diverse probe set.  To mitigate spurious biases, we retrained the model to be invariant to detected prediction shortcuts using a novel regularized training objective. We theoretically demonstrated that this objective effectively unlearns the spurious correlations captured during training. Experiments confirmed that our method successfully mitigates spurious biases and enhances model robustness to distribution shifts. Future work may explore constructing a more diverse probe set to further enhance the detection and mitigation of spurious biases.
 \section*{Acknowledgments}
This work is supported in part by the US National Science Foundation under grants 2217071, 2213700, 2106913, 2008208.
\bibliographystyle{named}

\bibliography{ijcai25}

\appendix
 \clearpage
\appendix
\section{Appendix}

\subsection{Proof of Proposition 1}
The proposition uses the result from Lemma 1 in our main paper, which is a restatement of Eq. (4.4) in \cite{Bombari2024how}. To prove the proposition, we first show that $\mathbf{O}\mathbf{O}=\mathbf{O}$. Note that $\mathbf{O}=\mathbf{I}-\mathbf{V}^T(\mathbf{V}\mathbf{V}^T)^{-1}\mathbf{V}$, thus we have
\begin{align}
    \mathbf{O}\mathbf{O}&= \big(\mathbf{I}-\mathbf{V}^T(\mathbf{V}\mathbf{V}^T)^{-1}\mathbf{V}\big)\big(\mathbf{I}-\mathbf{V}^T(\mathbf{V}\mathbf{V}^T)^{-1}\mathbf{V}\big)\nonumber\\
    &=\mathbf{I}-2\mathbf{V}^T(\mathbf{V}\mathbf{V}^T)^{-1}\mathbf{V}+\mathbf{V}^T(\mathbf{V}\mathbf{V}^T)^{-1}\mathbf{V}\nonumber\\
    &=\mathbf{I}-\mathbf{V}^T(\mathbf{V}\mathbf{V}^T)^{-1}\mathbf{V}=\mathbf{O}.
\end{align}
Next, we expand $\varphi(\tilde{x})^T\mathbf{O}\varphi(x)$ as follows
\begin{align}
    \varphi(\tilde{x})^T\mathbf{O}\varphi(x) &= \varphi(\tilde{x})^T\mathbf{O}\mathbf{O}\varphi(x)\\
    &= (\mathbf{O}\varphi(\tilde{x}))^T(\mathbf{O}\varphi(x))\label{eq:convert-to-inner-product}\\
    &\leq \|\mathbf{O}\varphi(\tilde{x})\|_2\|\mathbf{O}\varphi(x)\|_2 \label{eq:convert-to-norm},
\end{align}
where Eq.~\eqref{eq:convert-to-inner-product} uses the fact that $\mathbf{O}^T=\mathbf{O}$, and \eqref{eq:convert-to-norm} is the result of Cauchy–Schwarz inequality. 

The inequality \eqref{eq:convert-to-norm} holds in general. However, the equality actually holds in our setting. 
To show this, we need to prove that the vectors $\mathbf{O}\varphi(\tilde{x})$ and  $\mathbf{O}\varphi(x)$ are independent. We first note that a spurious sample $\tilde{x}$ is independent of an original sample $x$. For example, in the Waterbirds dataset \cite{sagawa2019distributionally}, let $\tilde{x}$ represent an image showing only a water background, and $\tilde{x}$ is independent of $x$, as $\tilde{x}$ may be obtained by removing core objects from images of landbirds or waterbirds with waterbird backgrounds. Thus, the corresponding feature vectors $\varphi(x)$ and $\varphi(\tilde{x})$ are  independent. If we assume that $\mathbf{O}\varphi(x)$ and $\mathbf{O}\varphi(\tilde{x})$ are dependent with $\mathbf{O}\varphi(x)=\eta\mathbf{O}\varphi(\tilde{x})$, where $\eta$ is a non-zero constant, then we have $\varphi(x)=\eta\mathbf{O}^{-1}\mathbf{O}\varphi(\tilde{x})=\eta\varphi(\tilde{x})$, which contradicts the fact that $\varphi(x)$ and $\varphi(\tilde{x})$ are independent. Therefore, $\mathbf{O}\varphi(x)$ and $\mathbf{O}\varphi(\tilde{x})$ are independent. Consequently, we have the following equality,
\begin{equation}
    \varphi(\tilde{x})^T\mathbf{O}\varphi(x) = \|\mathbf{O}\varphi(\tilde{x})\|_2\|\mathbf{O}\varphi(x)\|_2.
\end{equation}

Finally, we reinterpret the feature alignment $\gamma_\varphi$ as follows,
\begin{align}
\gamma_{\varphi}&=\mathbb{E}_{\tilde{x},x}\Big[\frac{\varphi(\tilde{x})^T\mathbf{O}\varphi(x)}{\|\mathbf{O}\varphi(x)\|_2^2}\Big]
\\
&=\mathbb{E}_{\tilde{x},x}\Big[\frac{\|\mathbf{O}\varphi(\tilde{x})\|_2\cdot \|\mathbf{O}\varphi(x)\|_2}{\|\mathbf{O}\varphi(x)\|_2\cdot \|\mathbf{O}\varphi(x)\|_2}\Big]\\
&=\frac{\mathbb{E}_{\tilde{x}}[\|\mathbf{O}\varphi(\tilde{x})\|_2]}{\mathbb{E}_{x}[\|\mathbf{O}\varphi(x)\|_2]}\label{eq:appendix-feature-alignment},
\end{align}
where Eq. \eqref{eq:appendix-feature-alignment} results from the fact that the random variables $x$ and $\tilde{x}$ are independent. 
\subsection{Learning algorithm}
We show the detailed training process of our proposed method, ShortcutProbe, in Algorithm~\ref{alg:1}. The algorithm is a two-step procedure. In the first step, we train a shortcut detector, and in the second step, we use the prediction shortcuts detected by the shortcut detector to mitigate spurious biases in the model.
\paragraph{Complexity analysis.} Given that the time complexity for obtaining $\mathcal{D}^y_{\text{cor}}$, $\mathcal{D}^y_{\text{pre}}$, and $\mathcal{D}^y_{\text{mis}}$ is $C_{\text{data}}$, the time complexity for each batch update during the shortcut detector training is $C_{\text{det}}$, and the time complexity for each batch update during classifier retraining is $C_{\text{ret}}$, the overall time complexity is $O(C_{\text{data}} + E_1N_BC_{\text{det}} + E_2N_BC_{\text{ret}})$.

Notably, the sets $\mathcal{D}^y_{\text{cor}}$, $\mathcal{D}^y_{\text{pre}}$, and $\mathcal{D}^y_{\text{mis}}$ can be precomputed before training and need to be constructed only once, allowing $C_{\text{data}}$ to be omitted once these sets are available. Additionally, $C_{\text{det}}$ and $C_{\text{ret}}$ are typically very small due to the lightweight design of the shortcut detector and the retraining process, which only involves the model’s final linear layer. Consequently, ShortcutProbe is highly computation-efficient. 
 We provide a run-time comparison between different debiasing methods in Table~\ref{tab:run-time} below.

\begin{table}[h]
\centering
\begin{tabular}{cccc}
\hline
JTT  & DFR & AFR & ShortcutProbe \\ \hline
1440 & 162 & 230 & 210           \\ \hline
\end{tabular}
\caption{Training time (s) comparison on the Waterbirds dataset.}\label{tab:run-time}
\end{table}

\begin{algorithm}[t]
		\caption{ShortcutProbe}
		$\mathbf{Input}$: Probe set $\mathcal{D}_{\text{prob}}$, parameters of an ERM-trained model $\theta$ including $\theta_1$ of the feature extractor and $\theta_2$ of the classifier, parameters of the shortcut detector $\psi=\mathbf{A}$ with $K$ base vectors, batch size $B$, number of batches per epoch $N_B$, number of training epochs for the first step $E_{1}$, learning rate $\alpha$ used in the first step,  number of training epochs for the second step $E_{2}$, learning rate $\beta$ used in the second step, regularization strengths $\eta$ and $\lambda$.\\
		$\mathbf{Output}$: the classifier's weights $\theta_2$
		\begin{algorithmic}[1]
\STATE{Obtain $\mathcal{D}^y_{\text{cor}}$, $\mathcal{D}^y_{\text{pre}}$, and $\mathcal{D}^y_{\text{mis}}$ for each class $y$ from $\mathcal{D}_{\text{prob}}$ using Eq.~\eqref{eq:correct-samples} and Eq.~\eqref{eq:predicted-samples}, respectively}
\STATE{//Learn shortcut detector}
        \FOR{$e=1,\ldots,E_1$}
        \FOR{$b=1,\ldots,N_B$}
        \STATE{Sample $\mathcal{B}^y_{\text{cor}}\subset\mathcal{D}^y_{\text{cor}}$ and $\mathcal{B}^y_{\text{pre}}\subset\mathcal{D}^y_{\text{pre}}$, $\forall y\in\mathcal{Y}$, with $|\mathcal{B}^y_{\text{cor}}|=|\mathcal{B}^y_{\text{pre}}|$ and $\sum_{y\in\mathcal{Y}}(|\mathcal{B}^y_{\text{pre}}|+|\mathcal{B}^y_{\text{cor}}|)=B$}
        \STATE{Calculate $\psi=\psi-\alpha\nabla_{\psi}(\mathcal{L}_{\text{det}}+\eta\mathcal{L}_{\text{reg}})$ using $\mathcal{B}^y_{\text{cor}}$ and $\mathcal{B}^y_{\text{pre}}$}
        \ENDFOR
        \ENDFOR
        \STATE{//Mitigate spurious biases}
        \FOR{$e=1,\cdots,E_2$} 
         \FOR{$b=1,\cdots,N_B$}
          \STATE{Sample $\mathcal{B}^y_{\text{cor}}\subset\mathcal{D}^y_{\text{cor}}$ and $\mathcal{B}^y_{\text{mis}}\subset\mathcal{D}^y_{\text{mis}}$, $\forall y\in\mathcal{Y}$, with $|\mathcal{B}^y_{\text{cor}}|=|\mathcal{B}^y_{\text{mis}}|$ and $\sum_{y\in\mathcal{Y}}(|\mathcal{B}^y_{\text{mis}}|+|\mathcal{B}^y_{\text{cor}}|)=B$}
        \STATE{Calculate $\theta_2=\theta_2-\beta\nabla_{\theta_2}\lambda\mathcal{L}_{\text{ori}}/\mathcal{L}_{\text{spu}}$ using $\mathcal{B}^y_{\text{cor}}$ and $\mathcal{B}^y_{\text{mis}}$}
        \ENDFOR
        \ENDFOR
        \RETURN $\theta_2$ 
		\end{algorithmic}\label{alg:1}
\end{algorithm}

\subsection{Datasets}

\begin{table*}[t]
\centering
\begin{tabular}{lclccc}
\toprule
\multicolumn{1}{l}{\multirow{2}{*}{Dataset}} & \multicolumn{1}{l}{\multirow{2}{*}{\begin{tabular}[c]{@{}c@{}}Number of \\ classes\end{tabular}}} & \multicolumn{1}{c}{\multirow{2}{*}{$\langle$class, attribute$\rangle$}} & \multicolumn{3}{c}{Number of images} \\ \cmidrule{4-6} 
\multicolumn{1}{l}{}                         & \multicolumn{1}{l}{}                                   & \multicolumn{1}{l}{}                                                  & Train       & Val        & Test      \\ \midrule
\multirow{4}{*}{Waterbirds \cite{sagawa2019distributionally}   }               & \multirow{4}{*}{2}                                     & $\langle$landbird, land$\rangle$                                        & 3,498       & 467        & 2,255     \\
                                             &                                                        & $\langle$landbird, water$\rangle$                                       & 184         & 466        & 2,255     \\
                                             &                                                        & $\langle$waterbird, land$\rangle$                                       & 56          & 133        & 642       \\
                                             &                                                        & $\langle$waterbird, water$\rangle$                                      & 1,057       & 133        & 642       \\ \midrule
\multirow{4}{*}{CelebA \cite{liu2015deep}}                      & \multirow{4}{*}{2}                                     & $\langle$non-blond, female$\rangle$                                     & 71,629      & 8,535      & 9,767     \\
                                             &                                                        & $\langle$non-blond, male$\rangle$                                       & 66,874      & 8,276      & 7,535     \\
                                             &                                                        & $\langle$blond, female$\rangle$                                         & 22,880      & 2,874      & 2,480     \\
                                             &                                                        & $\langle$blond, male$\rangle$                                           & 1,387       & 182        & 180       \\ \midrule
NICO    \cite{he2021towards}                                      & 10                                                     & $\langle$object, context  $\rangle$                                                                   & 10298        & 642       & 894      \\ \midrule
ImageNet-9   \cite{ilyas2019adversarial}                                & 9                                                      & N/A                                                                     & 54,600      & 2,100      & N/A         \\ 
ImageNet-A \cite{hendrycks2021natural}                                  & 9                                                      & N/A                                                                     & N/A           & N/A         & 1087      \\\midrule
CheXpert   \cite{irvin2019chexpert}                               & 2            &   $\langle$diagnose, race+gender  $\rangle$                                                                   & 167093          & 22280          & 33419      \\ \midrule
\multirow{6}{*}{MultiNLI \cite{williams2017broad}}                              & \multirow{6}{*}{3}           &   $\langle$contradiction, no negation$\rangle$                                                                   & 57498         &  22814          &  34597      \\
& & $\langle$contradiction, negation$\rangle$ & 11158 & 4634 & 6655\\
& & $\langle$entailment , no negation$\rangle$ & 67376 & 26949 & 40496\\
& & $\langle$entailment, negation$\rangle$ & 1521 &613 &886\\
& & $\langle$neither, no negation$\rangle$ & 66630 & 26655 & 39930\\
& & $\langle$neither, negation$\rangle$ & 1992 & 797 & 1148\\\midrule
\multirow{4}{*}{CivilComments \cite{borkan2019nuanced}}                              & \multirow{4}{*}{2}           &   $\langle$neutral , no identity$\rangle$     &     148186 &25159 & 74780     \\
& & $\langle$neutral , identity$\rangle$  & 90337 & 14966 & 43778 \\
& & $\langle$toxic , no identity$\rangle$  & 12731 & 2111 & 6455 \\
& & $\langle$toxic , identity$\rangle$  & 17784 & 2944&  8769  \\
\bottomrule
\end{tabular}%
\caption{Detailed statistics of the 8 datasets. $\langle$class, attribute$\rangle$ represents a spurious correlation between a class and a spurious attribute. ``N/A" denotes not applicable.}\label{tab:dataset-statistics}
\end{table*}

Table \ref{tab:dataset-statistics} gives detailed statistics for all the eight datasets.  We give the number of training, validation, and test images in each group specified by classes and attributes for the Waterbirds, CelebA, MultiNLI, and CivilComments datasets. For example, the group label $\langle$landbird, land$\rangle$ in the Waterbirds dataset has 3498 training images which are all landbirds and have land backgrounds.

\begin{table}[ht]
\centering
\small
\begin{tabular}{ccc}
\toprule
\multirow{2}{*}{Class}    & \multicolumn{2}{c}{Contexts}  \\ \cmidrule{2-3}
          &   Validation   & Test\\ \midrule
dog      & running & in\_street        \\
cat      & on\_tree & in\_street     \\
bear     & on\_tree & white          \\
bird     & on\_shoulder & in\_hand  \\
cow      & spotted & standing    \\
elephant & in\_circus & in street      \\
horse    & running & in\_street \\
monkey   & climbing & sitting       \\
rat      & running & in\_hole         \\
sheep    & at\_sunset & on\_road    \\ 
\bottomrule
\end{tabular}%
\caption{Classes and their associated contexts in the NICO datasets. Contexts not shown in the table are used in the training set.}\label{tab:class-context-nico}
\end{table}

NICO \cite{he2021towards} is a real-world dataset
for evaluating a method's out-of-distribution generalization performance.  NICO provides context labels as spurious attributes.   We used its Animal subset containing 10 object classes and 33 context labels. Following the setting in \cite{bai2021decaug,tiwari2023overcoming}, we  split the dataset into training, validation, and test sets with each set having unique contexts.   Table \ref{tab:class-context-nico} gives the allocation of the contexts for the 10 classes. 

ImageNet-9 \cite{bahng2020learning} is a subset of ImageNet, and ImageNet-A contains real-world images
that are challenging to the image classifiers trained on standard ImageNet. Both datasets do not have clear group partitions specified by the class and attribute associations.
 ImageNet-9 has 9 super-classes, i.e., Dog, Cat, Frog, Turtle, Bird, Primate, Fish, Crab, Insect, obtained by merging similar classes from ImageNet.   We extract images of the 9 super-classes from the ImageNet-A dataset and use these images for testing.

 The CheXpert dataset \cite{irvin2019chexpert} is a chest X-ray dataset from the Stanford University Medical center. There are six spurious attributes in the dataset, each of them is the combination of race (White, Black, Other) and gender (Male, Female). Two diagnose results, i.e., ``No Finding" (positive) and ``Finding" (negative) are the labels.

\subsection{Training Details}
\paragraph{ERM training.} This step trains ERM models which serve as the base models used in our framework for detecting prediction shortcuts and mitigating spurious biases. The training hyperparameters as well as the optimizer and learning rate scheduler used for each dataset are given in Table~\ref{table:erm-train}. For vision models, we initialized them with ImageNet-pretrained weights. For text models, we initialized them with weights pretrained on Book Corpus and English Wikipedia data.

\begin{table*}[t]
\begin{tabular}{cccccccc}
\toprule
Dataset       & Batch size & Epochs & Initial learning rate & Weight decay & Learning rate scheduler & Optimizer \\ \midrule
Waterbirds    & 32         & 100    & 0.003         & 0.0001       & CosineAnnealing         & SGD       \\
CelebA        & 128        & 20     & 0.003         & 0.0001       & CosineAnnealing         & SGD       \\
CheXpert      & 128        & 20     & 0.003         & 0.0001       & CosineAnnealing         & SGD       \\
MultiNLI      & 16         & 10     & 0.00001       & 0.0001       & Linear                  & AdamW     \\
CivilComments & 16         & 10     & 0.00001         & 0.0001       & Linear                  & AdamW     \\
NICO          & 128        & 100    & 0.003         & 0.0001       & CosineAnnealing         & SGD       \\
ImageNet-9    & 128        & 100    & 0.003         & 0.0001       & CosineAnnealing         & SGD       \\ \bottomrule
\end{tabular}
\caption{Training settings for training ERM models on different datasets.}\label{table:erm-train}
\end{table*}

\paragraph{Training ShortcutProbe models.} We provide hyperparameter settings for the experiments on the Waterbirds, CelebA, CheXpert, MultiNLI, CivilComments, ImageNet-9, and NICO datasets in Table~\ref{tab:hyper-setting}. We used an SGD optimizer with a momentum of 0.9 and a weight decay of $1\times10^{-4}$ in training the shortcut detector and retraining the classifier. We chose $K$ from $\{2,4, 6, 8\}$, $\eta$ from $\{0.1,1.0,5.0,10.0\}$, $\lambda$ from $\{0.1,1.0,5.0,10.0,50.0\}$, and $N_B$ from $\{50,100,200\}$, $\beta$ from $\{0.0001,0.0005,0.001,0.003,0.01\}$, and $r$ from $\{0.1,0.2,0.3,0.4,0.5\}$. We selected the best hyperparameters based on the performance on the whole validation set if the training data was used to construct the probe set or the remaining validation set if part of the validation set was used for the construction. The remaining hyperparameters were determined based on our empirical observations considering both dataset size and the convergence of training.

\paragraph{Training baseline models.} For JTT \cite{liu2021just}, we combined the training data with half of the validation data to create a new training set for training JTT models. For DFR \cite{kirichenko2022last} and AFR \cite{qiu2023simple}, we applied these methods to the same ERM-trained model to ensure a fair comparison. We adhered to the hyperparameter settings recommended in the respective original papers.

\begin{table*}[ht]
\centering
\begin{tabular}{ccccccccccc}
\toprule
Dataset    & $K$ & $\eta$ & $\lambda$ & $E_1$ & $E_2$ & $B$ & $N_B$ & $\alpha$ & $\beta$ & $r$ \\ \midrule
Waterbirds & 2   & 5.0  & 5.0     & 50    & 50    & 32 & 200 & 0.0001   & 0.001  & 0.3      \\
CelebA     & 2   & 5.0  & 5.0     & 50   & 50    & 128 & 100 & 0.0001   & 0.001  & 0.1       \\
CheXpert & 6   & 10.0  & 50.0     & 50   & 50    & 128 & 50 & 0.0001   & 0.003  & 0.1       \\
MultiNLI & 2   & 5.0  & 5.0     & 50  & 50    & 128 & 100 & 0.0001   & 0.001  & 0.1       \\
CivilComments & 6   & 1.0  & 1.0     & 50   & 50    & 128 & 100 & 0.0001   & 0.003  & 0.1       \\
NICO       & 8   & 1.0  & 1.0    & 50  & 50    & 128 & 200 & 0.0001   & 0.001&  -      \\ 
ImageNet-9 & 4   & 1.0   & 1.0     & 50   & 50    & 128 & 200 & 0.0001   & 0.001  & -     \\ \bottomrule
\end{tabular}%
\caption{Hyperparameter settings for experiments on the seven datasets. $K$: number of base vectors; $\eta$: regularization strength for the semantic similarity constraint in Eq.~\eqref{eq:shortcut-detector-reg}; $\lambda$: regularization strength used in the training objective in Eq.~\eqref{eq:overall-objective}; $E_1$: number of training epochs for learning the shortcut detector; $E_2$: number of training epochs for retraining the classifier; $B$: batch size; $N_B$: number of batches sampled in each epoch; $\alpha$: learning rate for learning the shortcut detector; $\beta$: learning rate for retraining the classifier; $r$: proportion of samples used to construct the probe set. When $r$ is not specified (``-"), it means using the training data to construct the probe set.}\label{tab:hyper-setting}
\end{table*}

\section{Additional results}
\paragraph{ImageNet-9 and ImageNet-A.} We presents performance comparison on the ImageNet-9 and ImageNet-A datasets in Table \ref{tab:imagenet}. The validation accuracy measures the in-distribution performance of a model, while the accuracy gap measures the performance drop from ImageNet-9 to ImageNet-A. Images in the ImageNet-A dataset represent distribution shifts from the training images in the ImageNet-9 dataset. Thus, the accuracy on the ImageNet-A dataset measures a model’s performance under distribution shifts. As shown in Table \ref{tab:imagenet}, our method achieves the best on the ImageNet-A dataset, demonstrating its robustness to distribution shifts. It also exhibits a good tradeoff between in-distribution and out-of-distribution performance by achieving the best accuracy gap.

\paragraph{ResNet-152 and ViT backbones.} Our method can be easily applied to larger backbone networks beyond ResNet-50, such as ResNet-152 and ViT. We evaluated our method with ResNet-152 and ViT-B/32 on three vision datasets and provide a performance comparison with baseline methods in Table~\ref{tab:performance-comparison} below. We observe that our method remains highly effective on large-scale models.

\begin{table}[h]
\resizebox{\columnwidth}{!}{%
\begin{tabular}{ccccc}
\hline
Backbone                    & Method & Waterbirds & CelebA & Chexpert \\ \hline
\multirow{4}{*}{ResNet-152} & ERM    & 17.4       &   60.6     &  18.6        \\
                            & DFR    & 30.3       &  68.3      &  66.0        \\
                            & AFR    & 31.4       &  70.7      &       63.8   \\
                            & Ours   & \textbf{34.7}       &   \textbf{80.0}     &   \textbf{68.3}       \\ \hline
\multirow{4}{*}{ViT-B/32}        & ERM    & 69.5       & 52.2   & 20.9     \\
                            & DFR    & 87.6       & 63.0   & 73.4     \\
                            & AFR    & 86.6       & 79.5   &   64.5       \\
                            & Ours   & \textbf{88.0}       & \textbf{83.7}   & \textbf{75.1}     \\ \hline
\end{tabular}
}
\caption{Comparison of worst-group accuracy (\%) across last-layer retraining methods using ResNet-152 and ViT backbones.}\label{tab:performance-comparison}
\end{table}

\begin{table*}[t]
\centering
\resizebox{0.6\linewidth}{!}{%
\begin{tabular}{p{5cm}@{\hskip 0.0cm}>{\centering}ccc}
\toprule
Method                          & ImageNet-9  ($\uparrow$)  & ImageNet-A  ($\uparrow$) & Acc. gap ($\downarrow$) \\ \midrule
ERM                             & 90.8$_{\pm0.6}$  & 24.9$_{\pm1.1}$ & 65.9     \\
ReBias \cite{bahng2020learning} & 91.9$_{\pm1.7}$  & 29.6$_{\pm1.6}$ & 62.3     \\
LfF \cite{nam2020learning}      & 86.0          & 24.6         & 61.4     \\
CaaM \cite{wang2021causal}      & 95.7          & 32.8         & 62.9     \\
SSL+ERM \cite{kim2022learning}  & 94.2$_{\pm0.1}$  & 34.2$_{\pm0.5}$ & 60       \\
LWBC\cite{kim2022learning}      & 94.0$_{\pm0.2}$  & 36.0$_{\pm0.5}$ & 58       \\
SIFER \cite{pmlrv202tiwari23a}  & \textbf{97.8}$_{\pm0.1}$ & 40.0$_{\pm0.8}$ & 57.8     \\ 
\cellcolor[HTML]{EFEFEF} \textbf{ShortcutProbe (Ours)}            &\cellcolor[HTML]{EFEFEF} 96.9$_{\pm0.2}$          &\cellcolor[HTML]{EFEFEF} \textbf{45.3}$_{\pm1.2}$         & \cellcolor[HTML]{EFEFEF}\textbf{51.6}    \\ \bottomrule
\end{tabular}%
}
\caption{Comparison of average accuracy (\%) and accuracy gap (\%) on the ImageNet-9 and ImageNet-A datasets.}\label{tab:imagenet}
\end{table*}
\section{Qualitative Analysis of Learned Prediction Shortcuts}
\begin{figure}[ht]
    \centering
    \includegraphics[width=0.9\linewidth]{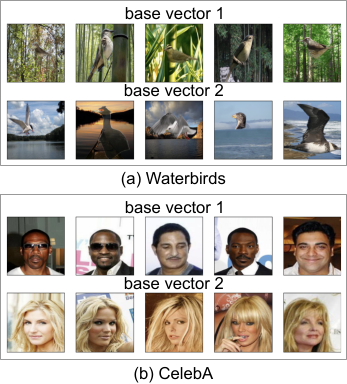}
    \caption{Visualization of the top-5 images that are most similar to the learned base vectors from the (a) Waterbirds and (b) CelebA datasets.}
    \label{fig:spurious-attributes}
\end{figure}

To qualitatively analyze the detected prediction shortcuts, we aim to interpret the learned base vectors in the matrix $\mathbf{A}$, as prediction shortcuts are obtained by linearly combining these vectors. To achieve this, for each base vector, we gave the top-5 images whose prediction shortcuts are most similar to the base vector. Specifically, for each learned base vector, we first calculated the embeddings of training samples, and following Eq.~\eqref{eq:projection-matrix}, we extracted prediction shortcuts in those samples as projected vectors by projecting the embeddings to the subspace spanned by the learned base vectors. We ranked the images based on the similarity of their prediction shortcuts to the base vector. A large similarity score  signals a strong existence of the feature the base vector represents in the corresponding image. 

As shown in Fig. \ref{fig:spurious-attributes}(a), on the Waterbirds dataset, the two learned base vectors are most similar to images with land backgrounds and water backgrounds, respectively. In Fig. \ref{fig:spurious-attributes}(b), the two learned base vectors are most similar to images of male celebrities and female celebrities, respectively. These results show that our shortcut detector can learn spurious attributes that well align with the biases in the datasets which models tend to capture during training.
\end{document}